# ViTCN: Vision Transformer Contrastive Network For Reasoning

Bo Song*

Khoury College of Computer Science, Northeastern University

Boston, USA

song.bo1@northeastern.edu

Yuanhao Xu

Khoury College of Computer Science, Northeastern University

Boston,USA

xu.yuanh@northeastern.edu

Yichao Wu

Khoury College of Computer Science, Northeastern University

Boston USA

wu.yicha@northeastern.edu

*Abstract*—Machine learning models have achieved significant milestones in various domains, for example, computer vision models have an exceptional result in object recognition, and in natural language processing, where Large Language Models (LLM) like GPT can start a conversation with human-like proficiency. However, abstract reasoning remains a challenge for these models, "Can AI really thinking like a human?" still be a question yet to be answered. Raven's Progressive Matrices (RPM) is a metric designed to assess human reasoning capabilities. It presents a series of eight images as a problem set, where the participant should try to discover the underlying rules among these images and select the most appropriate image from eight possible options that best completes the sequence. This task always be used to test human reasoning abilities and IQ. Zhang et al proposed a dataset called RAVEN [12] which can be used to test Machine Learning model abstract reasoning ability. In this paper, we purposed Vision Transformer Contrastive Network which build on previous work with the Contrastive Perceptual Inference network (CoPiNet), which set a new benchmark for permutation-invariant models Raven's Progressive Matrices by incorporating" contrast effects" from psychology, cognition, and education, and extends this foundation by leveraging the cutting-edge Vision Transformer architecture. This integration aims to further refine the machine's ability to process and reason about spatial-temporal information from pixel-level inputs and global wise features on RAVEN dataset.

## I. INTRODUCTION

Computer Vision Reasoning has became a popular field within computer vision tasks, with the development of machine learning models like convolutional neural networks[1] and ResNet[2] demonstrating remarkable capabilities in feature extraction and classification prediction. However, the question of whether these models can truly" think in pictures" remains a significant challenge. Raven's Progressive Matrices (RPM) is one of the most widely used metrics for assessing human reasoning capabilities. To evaluate the reasoning abilities of machine models, RPM has been employed as a critical metric for evaluation. RPM contains a 3X3 matrix of visual patterns where the final piece is missing, and one of the 8 images from answer set should be selected and complete matrix based on observed underlying rules.

In previous research, researchers used CNN[1] and ResNet[2] on RAVEN dataset, and the result still has a huge performance gap between human performance. Zhang et al proposed a new architecture called CopiNet[3], which uses a CNN as an encoder and constructs a contrast network to make a prediction, CoPinet incorporates a novel contrast module and contrast loss to improve machine reasoning. It is designed to be permutation-invariant, addressing the challenge that RPM poses due to its requirement for both perception and inference capabilities in solving problems. The introduction of a contrast mechanism, based on perceptual learning and educational psychology, aims to enhance the model's ability to generalize from contrastive data presentations.

In our proposed module, we are using Vision Transformers to substitute the CNN encoder, Vision Transformer is purposed by Kolesnikov et al [4], compare to CNN which perform a computation-intensive task of extracting local features from images, Vision Transformer convert an image into a sequence of non-overlapping patches, each patch is then transformed into a vector, which can grasping global information within images, transcending the limitations of local feature extraction.

We have three major contributions in this work:

- We introduce a new contrastive vision reasoning network, an approach designed to enhance machine reasoning capabilities by combining contrastive learning principles with vision transformer.

- Our model achieves a notable improvement on the RAVEN dataset, particularly excelling in center RPM problems. We report a 1.73% overall performance increase over the current state-of-the-art.

- By combining contrast reasoning networks with the vision transformer architecture, our model emphasizes global feature extraction over local features. This integration significantly enhances the robustness of the vision reasoning model.

## II. RELATED WORK

### A. Transformer

Following their success in NLP, transformers have been adapted to computer vision tasks, leading to the development of Vision Transformers (ViT). ViT, introduced by Dosovitskiyet al [4], applies the transformer architecture directly to sequences of image patches, treating them similarly to tokens (words) in a sentence. This approach has demonstrated competitive performance in image classification tasks, challenging the long-standing dominance of convolutional neural networks (CNNs) in the field.

The rise of transformer models has transformed areas in intelligence, especially in natural language processing (NLP) and more recently in computer vision. Transformers, as introduced by Vaswani et al [4. In the paper" Attention is All You Need" are built on self attention mechanisms that enable models to assign levels of importance to various parts of the input data. This capability has brought progress in comprehending and generating language with models, like BERT (Bidirectional Encoder Representations from Transformers) [5] and GPT (Generative Pretrained Transformer) [6] achieving remarkable outcomes in tasks such as language translation question answering and text generation.

Building on their accomplishments in NLP transformers have been modified for computer vision tasks resulting in the creation of Vision Transformers (ViT)[4]. ViT, presented by Dosovitskiy et al. employs the architecture directly to sequences of image patches treating them akin to tokens (words) in a sentence. This method has showcased performance in image classification assignments challenging the dominance of convolutional neural networks (CNNs), within the field.

### B. Visual Reasoning on RPM

From the inception of ImageNet [1], researchers have utilized large deep learning networks alongside huge datasets to train machine learning models for vision pattern recognition tasks. Hoshen & Werman [7] first expanded the application of deep learning models to IQ questions, training a CNN model but it did not surpass human performance. Santoro et al. [8] experimented with more deep learning models on reasoning tasks, such as LSTM [9] and ResNet [2], but still did not achieve optimal performance. It appears that pure deep learning networks may not excel in reasoning tasks, prompting researchers to incorporate auxiliary annotations into their models. These annotations represent logical rules and attributes of the RPM questions, may help generate more appropriate feature representations. Several models have achieved significant success by following this approach, including WReN [8], MXGNet [10], and ACL [11]. A major advancement was made by Zhang et al., who proposed an RPM dataset with structured annotation named RAVEN [12], and the Contrastive Perceptual Inference network (CoPiNet)[3]. CoPiNet introduced a contrastive module to assist the model in learning the underlying rules from given questions, achieving state-of-the-art results. After RAVEN dataset, I-RAVEN [14] and RAVEN-FAIR[15] are proposed to fix the defects of RAVEN. The new interests comes from probabilistic methods and algebraic methods in RPM[16,17]

## III. METHODS

In this study we make use of the RAVEN dataset by Zhang et al. [12] in which every set consists of 16 images. The first 8 images form a question while the following 8 images make up the answer set. As a result, the overall data input size is N × 16 × 96 × 96, with N representing the size. 96 indicating the dimensions of each image.

In this paper, we propose VitCN - Vision Transformer Contrastive Network which construct by Vision Transformer, contrasting, perceptual inference and permutation invariance. The feature encoder is constructed by vision transformer encoder that includes MLP and multi-head attention module. The contrast network contains two branches, the inference branch and the perception branch. The inference branch samples a most liked rule and feed back to perception to make the final prediction.

The Vision Transformer (ViT) uses self attention mechanisms to analyze images by dividing them into patches. This approach allows ViT to preprocess input images for the RPM problem, creating feature maps for each patch. These feature maps act as in-depth representations of the images in the RPM matrix offering a foundation for the contrast module to utilize. Leveraging ViT as its core, ViTCN gains from its capacity to understand contexts within images establishing a groundwork, for future reasoning and inference activities.

For every answer candidate, the model begins by extracting features that represent the information it contains. Next the contrast module combines the shared features, from all answers. This combination involves summing or averaging the features found across the candidate set. The goal of this stage is to recognize the common features among the candidates providing a basis for contrasting. The module computes the contrast for each candidate by subtracting their features from the shared features. This process emphasizes the distinguishing features of each candidate in comparison to the average. By concentrating on these features, the model is better able to make decisions based on each candidates unique features. This method mirrors how humans typically approach problem solving by comparing and contrasting choices to determine which one is most appropriate.

Initially, the input is fed into a Vision Transformer encoder, which splits each image into several patches, flattens these patches, and performs a linear projection to obtain patch embedding, P is denoted as patch size, an input with shape (N, C, W, W) will output a token matrix with size (N, $(W/P)^2$, $P^2$)

$$\mathbf{z}_0 = \left[ \mathbf{x}_{\text{class}} ; \mathbf{x}_p^1 \mathbf{E} ; \mathbf{x}_p^2 \mathbf{E} ; \cdots ; \mathbf{x}_p^N \mathbf{E} \right] + \mathbf{E}_{\text{pos}},$$
$$\mathbf{E} \in \mathbb{R}^{(P^2 \cdot C) \times D}, \mathbf{E}_{pos} \in \mathbb{R}^{(N+1) \times D} \quad (1) [4]$$

The Transformer Encoder is constructed with layers that include Multi-Headed Self-Attention modules and Multi-Layer Perceptron modules. Each layer begins with LayerNorm, followed by the respective module, and concludes with a residual connection. Then the matrix of token will be flatten to 1D token and feed into contrastive network. As shown in equation 1, 2, 3.

$$z'_\ell = \text{MSA}(\text{LN}(z_{\ell-1})) + z_{\ell-1}, \quad \ell = 1...L$$

(2) [4]

$$\mathcal{F}_{\mathcal{O} \cup a} = z_\ell = \text{MLP}(\text{LN}(z'_\ell)) + z'_\ell, \quad \ell = 1...L$$

(3) [4]

For module level contrast, we use following equations proposed by Zhang et al [3], where F denotes features from transformer and h is a combination of BatchNorm and Conv layer. The inference branch and perception branch share the same input, the inference branch are responsible for predicting the underlying rules and perception branch join with inference branch to make the final prediction. The module computes the contrast for each candidate by subtracting the aggregated common features from the individual features of each candidate. This operation highlights the features that are unique or distinguishing for each candidate compared to the collective average.

$$\text{Contrast}(\mathcal{F}_{\mathcal{O} \cup a}) = \mathcal{F}_{\mathcal{O} \cup a} - h\left(\sum_{a' \in \mathcal{A}} \mathcal{F}_{\mathcal{O} \cup a'}\right)$$

(4) [3]

$$\ell = \log(\sigma(f(\mathcal{O} \cup a_\star) - b(\mathcal{O} \cup a_\star))) + \sum_{a' \in \mathcal{A}, a' \neq a_\star} \log(1 - \sigma(f(\mathcal{O} \cup a') - b(\mathcal{O} \cup a')))$$

(5) [3]

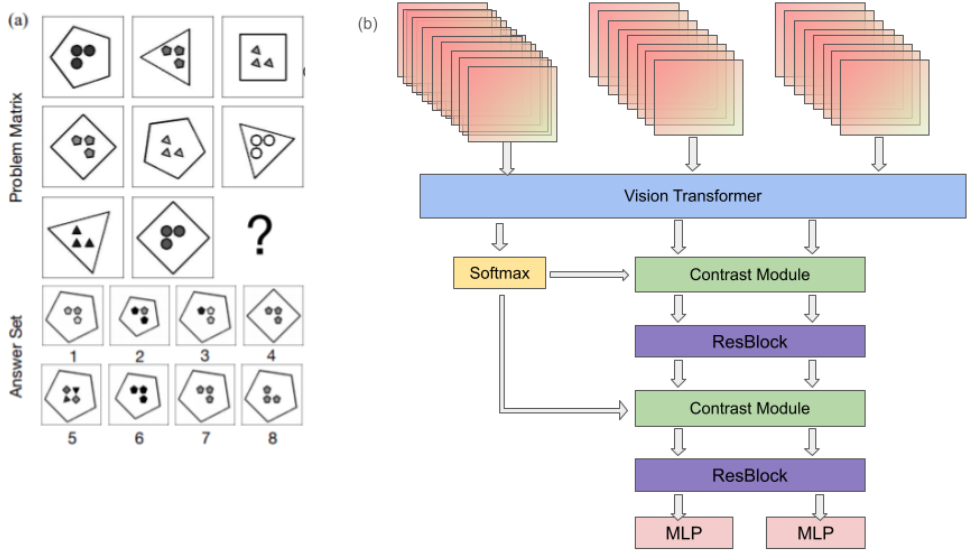

Figure 1. Figure a is an example of RAVEN dataset, in the RAVEN dataset there are 8 images forming a problem matrix followed by another set of 8 images representing the answers. This particular example demonstrates the underlying patterns related to variations in shape and color. As such all 9 images should consist of 3 circles, 3 triangles and 3 pentagons positioned both inside and outside with an arrangement. Moreover the internal shapes should exhibit three colors; white, grey and black. Therefore the correct choice is option number 7. Figure b showcases an overview of our proposed ViTCN.

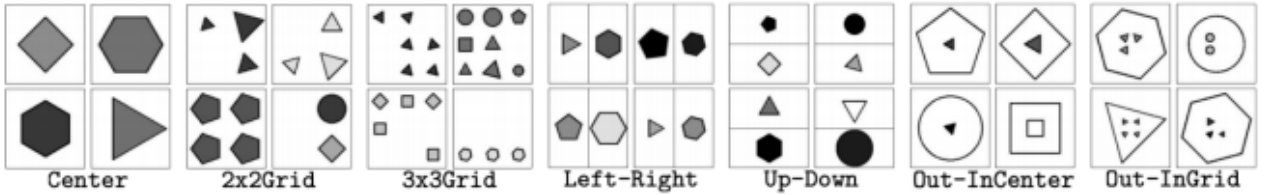

Figure 2. Example of 7 different figure configurations in RAVEN

## IV. EXPERIMENT

### A. Dataset and Experiment setup

In our research we make use of the RAVEN dataset, [12] a benchmark designed to test abstract visual reasoning capabilities. This dataset consists of 70,000 problems, each featuring a grid of 16 images arranged in a 3x3 layout with one image missing and an answer set containing 8 images. The challenge involves selecting the missing image from an answerset of 8 options based on underlying rules. The dataset covers seven types of figure configurations distributed evenly, offering a thorough assessment of spatial and logical reasoning abilities. Each sample's input size is denoted as $N \times 16 \times 96 \times 96$ where $N$ represents the size and $96 \times 96$ indicates the resolution of each image. While the original RAVEN dataset contains images, with a resolution of $160 \times 160$ pixels we resize all images to $96 \times 96$ pixels for efficiency and to match the input requirements of our Vision Transformer.

We use PyTorch to construct and finetune our model, use ADAMW [13] for model optimization, we divide the RAVEN dataset into training, validation, and testing sets following the standard distribution: 60% for training, 20% for validation, and 20% for testing. Our models are implemented in PyTorch and optimized using the ADAMW optimizer [13] with a learning rate of $1e-4$, reduced by half every 20 epochs based on the validation set performance.

We employ cross-entropy loss as our training objective. Training is performed with early stopping based on validation loss to prevent overfitting.

Our evaluation metric is accuracy, calculated as the percentage of correctly answered questions in the test set. We compare ViTCN's performance against state-of-the-art models on the RAVEN dataset, including traditional CNNs [1], LSTM [9], ResNet [2], WReN [8] and the original CoPINet, to demonstrate the efficacy of our contrastive reasoning mechanism.

TABLE I. TEST ACCURACY OF DIFFERENT MODELS ON RAVEN

| Model | Acc | Center | 2x2Grid | 3x3Grid | L-R | U-D | O-IC | O-IG |
|---|---|---|---|---|---|---|---|---|
| LSTM | 13.07% | 13.19% | 14.13% | 13.69% | 12.84% | 12.35% | 12.15% | 12.99% |
| WReN | 17.62% | 17.66% | 29.02% | 34.67% | 7.69% | 7.89% | 12.30% | 13.94% |
| CNN | 36.97% | 33.58% | 30.30% | 33.53% | 39.43% | 41.26% | 43.20% | 37.54% |
| ResNet | 53.43% | 52.82% | 41.86% | 44.29% | 58.77% | 60.16% | 63.19% | 53.12% |
| CoPiNet | 91.42% | 95.05% | 77.45% | 78.85% | 99.10% | **99.65%** | **98.50%** | **91.35%** |
| ViTCN(Our model) | **93.15%** | **97.02%** | **85.56%** | **85.40%** | **99.12%** | 99.21% | 98.14% | 90.50% |
| Human | 84.41% | 95.45% | 81.82% | 79.55% | 86.3% | 81.81% | 86.36% | 81.81% |

### B. Results

In this section we present the results of our ViTCN models experiments on the RAVEN dataset. Compare its performance with baseline models like LSTM, WReN, CNN, ResNet, CoPiNet and human performance. According to Table ?? our ViTCN model achieves an accuracy of 93.15% on the RAVEN test set surpassing the previous top performing CoPiNet model with an accuracy of 91.42%. This outcome highlights the effectiveness of combining Vision Transformers with reasoning mechanisms for handling visual reasoning tasks. Noteworthy is that ViTCN not outperforms all machine learning models but also exceeds human performance which stands at 84.41%.

In comparison CoPiNet demonstrates performance in 'L R' 'U D' 'O IC' and 'O IG' categories with its accuracy in 'U D' at 99.65%. However, ViTCN showcases superior performance particularly showcasing its prowess, in intricate grid patterns seen in the '2x2Grid' and '3x3Grid' configurations.

## V. CONCLUSION AND DISCUSSION

The excellent performance of ViTCN can be credited to its design that utilizes the Vision Transformer for in depth feature extraction and incorporates a contrastive reasoning approach to boost problem solving capabilities. By capturing global wise features in images and using a contrast mechanism to highlight distinctive features ViTCN shows a significant improvement in reasoning accuracy.

Additionally, the findings also shed light on the difficulty that abstract reasoning presents for both humans and AI models. The variations in performance across types of problems demonstrate the challenge of recognizing and applying abstract rules in diverse contexts. ViTCNs success across these scenarios underscores its adaptability and broad applicability in tasks involving reasoning.

In summary the ViTCN model establishes a new benchmark on the RAVEN dataset surpassing existing models and human performance in visual reasoning tasks. Its success across problem categories showcases the effectiveness of combining Vision Transformers with contrastive reasoning module. Future research will explore enhancements to the models' structure and its utilization, in tests related to reasoning and intelligence aiming to advance the field of AI driven reasoning.